\documentclass[conference]{IEEEtran}
\usepackage{hyperref} 
\IEEEoverridecommandlockouts
\usepackage{cite}
\usepackage{amsmath,amssymb,amsfonts}
\usepackage{algorithmic}
\usepackage{textcomp}
\usepackage{xcolor}
\usepackage{graphicx}
\usepackage{soul}
\usepackage{float}
\usepackage{tabularx}
\def\BibTeX{{\rm B\kern-.05em{\sc i\kern-.025em b}\kern-.08em
    T\kern-.1667em\lower.7ex\hbox{E}\kern-.125emX}}

\begin{document}

\title{Classification of Natural Language Processing Techniques for Requirements Engineering}

\author{Liping~Zhao, Waad~Alhoshan, Alessio~Ferrari, Keletso~J.~Letsholo

\thanks{L. Zhao is with the University of Manchester, Manchester, UK}

\thanks{W. Alhoshan is with Al-Imam Mohammed ibn Saud Islamic University, Riyadh, Saudi Arabia}

\thanks{A. Ferrari is with Consiglio Nazionale delle Ricerche, CNR-ISTI, Pisa, Italy}

\thanks{K.J. Letsholo is with Higher Colleges of Technology, Abu Dhabi, UAE}
}


\maketitle

\begin{abstract}
Research in applying natural language processing (NLP) techniques to requirements engineering (RE) tasks spans more than 40 years, from initial efforts carried out in the 1980s to more recent attempts with machine learning (ML) and deep learning (DL) techniques. However, in spite of the progress, our recent survey shows that there is still a lack of systematic understanding and organization of commonly used NLP techniques in RE. 
We believe one hurdle facing the industry is lack of shared knowledge of NLP techniques and their usage in RE tasks. In this paper, we present our effort to synthesize and organize 57 most frequently used NLP techniques in RE. We classify these NLP techniques in two ways: first, by their NLP tasks in typical pipelines and second, by their linguist analysis levels. We believe these two ways of classification are complementary, contributing to a better understanding of the NLP techniques in RE and such understanding is crucial to the development of better NLP tools for RE.
\end{abstract}

\begin{IEEEkeywords}
Requirements Engineering (RE), Natural Language Processing (NLP), NLP Tasks, NLP Techniques
\end{IEEEkeywords}

\section{Introduction}
\label{Introduction}

Research in developing natural language processing (NLP) support for requirements engineering (RE), or NLP4RE for short, dates back to the early 1980s and has seen a continuous flow of contributions in the past 40 years~\cite{ryan1993role,sawyer2005shallow,berzins2007innovations,berry2012case}. However, in spite of huge improvements and advances in NLP in the last 20 years \cite{hirschberg2015advances,goth2016deep}, and great progress in NLP4RE research in the last 10 years, the uptake of NLP technologies in RE, and their industrial penetration, is still limited and fragmented\cite{dalpiaz2018natural,zhao2021natural}. Thus large gaps remain between NLP4RE research and its practical application~\cite{zhao2021natural}. 

A recent survey~\cite{zhao2021natural} cites insufficient industrial evaluation of NLP4RE research, the lack of shared RE-specific language resources, and the lack of technology know-how in NLP among the reasons for these gaps. As a first step to close these gaps, this paper aims to classify the NLP techniques commonly used in RE so that they are easy to understand. We believe that a better understanding of the NLP techniques in RE is not only crucial to the development of better NLP tools for RE, but also to their industrial adoption. In particular, the paper lays foundations for establishing a common terminology and vocabulary of the NLP techniques through the following contributions:

\begin{itemize}
\item We extract and synthesize 57 commonly used NLP techniques in RE for NLP4RE research and practice. 
\item We systematically classify these techniques in two ways: by their tasks typically performed in NLP pipelines and then by their linguistic analysis capability.  
\end{itemize}

The paper is organized as follows. Sect.~\ref{sec:background} provides a brief history of NLP for RE, as the background and motivation for this paper. Sect.~\ref{sec:method} describes how we extract and synthesize the common set of NLP techniques for RE. Sect.~\ref{sec:NLPTasks} and Sect.~\ref{sec:NLPLevels} present our classification of these techniques. Sect.~\ref{sec:conclusion} concludes the paper.

\section{Background: 40 Years of NLP4RE}
\label{sec:background}

As a background to this paper, we provide a brief history of NLP4RE. We first point to some notable contributions in RE that use traditional NLP techniques, and then outline the recent application of machine learning (ML) and deep learning (DL) in RE. However, the focus of this paper is on NLP techniques, not ML and DL techniques. 

\subsection{Traditional NLP for RE}

The relationship between NLP and RE is well established and widely discussed, with supporters and detractors~\cite{ryan1993role,dalpiaz2018natural,ferrari2017natural,berry2017evaluation}. Pioneering researchers in the field are Chen~\cite{chen1983} and Abbott~\cite{abbott1983program}, who, in the early 1980s, proposed using syntactic features of English sentences for database modeling and program design. Abbott’s approach was subsequently adapted to a program design tool by Berry \textit{et al.}~\cite{berry1987}. These works were mostly based on extracting relevant entities from the requirements text through simple syntactic rules, assuming that NL requirements were expressed in some constrained, predictable format, which, however, is rarely the case in practice~\cite{booch2008object}. 

After these pioneering works, the beginning of 1990s saw some serious attempts to develop NLP4RE tools, introducing techniques to account for the complexity and variety of NL. Two well-known NLP tools, \textit{findphrases} by Aguilera and Berry~\cite{aguilera1990} and OICSI by Rolland and Proix~\cite{rolland1992natural}, were the results of these efforts. Both tools were still oriented to the extraction task~\cite{zhao2021natural}, also referred as abstraction~\cite{berry2017evaluation}, and used lexical affinity and semantic cases, respectively, two techniques that are far more sophisticated than those previously used. 

For the remaining 1990s right up to the beginning of 2000s, a succession of NL tools had been proposed, among which were AbstFinder by Goldin and Berry~\cite{goldin1997abstfinder}, NL-OOPS by Mich~\cite{mich1996nl}, Circe by Ambriola and Gervasi~\cite{ambriola2006systematic}, CM-Builder by Harmain and Gaizauskas~\cite{harmain2000cm}. These works normally use traditional rule-based NLP  techniques, and are oriented to term extraction and model generation. Other tools, such as QuARS by Fabbrini \textit{et al.}~\cite{fabbrini2001linguistic}, and ARM by Wilson \textit{et al.}~\cite{wilson1997automated}, focus on defect detection and mostly use dictionary-based techniques. 

The early 2000s appeared to be a period of experimentation of new NLP techniques and new ideas addressing other tasks and phases of the RE process. Information retrieval (IR) techniques were used to improve requirements tracing~\cite{hayes2003improving}, statistical NLP techniques were applied to identify ``shallow knowledge'' from requirements text~\cite{sawyer2005shallow}, and to tracing relationships between requirements~\cite{cleland2007best}. 

Since the late 2000s, NLP4RE has become a full-fledged research area, attracting researchers from the wider RE community. A large number of tools have since been developed, among which are SREE (Tjong and Berry~\cite{tjong2013design}) for ambiguity detection and aToucan (Yue et al.~\cite{yue2015atoucan}) for model generation. Further developments include tools detection of defects~\cite{ferrari2018detecting}, smells~\cite{femmer2017rapid} and equivalent requirements~\cite{falessi2011empirical}.

Given the increasing need to make software systems trustworthy, accountable, legally compliant, as well as security- and privacy-aware, NLP has been largely applied also to legal documents~\cite{sleimi2018automated} and privacy policies~\cite{bhatia2016mining}, in the field of RE and Law. Finally, to support agile software development, requirements expressed in the form of user stories have been identified as an interesting area of application for NLP~\cite{robeer2016automated}. 

\subsection{Machine Learning and Deep Learning for RE}

Following the development of successful statistical NLP methods based on ML in the 1990s \cite{hirschberg2015advances,cambria2014jumping}, ML techniques have become increasingly important to NLP. The advantages of the ML-based approaches over the traditional, rule-based NLP approaches are effectiveness, considerable savings in terms of expert manpower, and straightforward portability to different domains \cite{sebastiani2002machine}. 

In RE, the earliest adoption of ML to NLP can be traced to a study by Cleland-Huang et al. \cite{cleland2007automated}, published in 2007, in which the authors presented an approach for automatically detecting and classifying non-functional requirements (NFRs) from requirements documents. The approach uses a set of weighted indicator terms to classify requirements; a probability value of each indicator term is computed by a probability function similar to Naïve Bayes \cite{lewis1998naive}, to estimate the likelihood of an input requirement being classified into a certain NFR category. The development of this approach thus marked the beginning of the work on ML-based approaches for RE and, as a seminal work in this area, this approach has been frequently used as the baseline to assess the performance of new techniques~\cite{kurtanovic2017automatically,dalpiaz2018natural}. 

With the recent widespread availability of NL content relevant to RE, such as feedback from users in app stores and social media, and developers’ comments in discussion forums and bug tracking systems, we have observed a rising interest in using ML techniques to support data-driven RE~\cite{maalej2015toward} and crowd-based RE~\cite{groen2017crowd}. These areas aim to leverage information available from stakeholders’ implicit and explicit feedback, including diverse sources as app reviews~\cite{maalej2016automatic}, issue tracking systems~\cite{merten2016software}, Twitter~\cite{guzman2017little} or user fora~\cite{morales2019speech}, to improve RE activities such as requirements elicitation and prioritization. Most of the works use ML techniques, as these can be effectively exploited when the task can be reduced to a classification problem, and a large amount of data is available. The analysis of different forms of feedback can be regarded as the main trend of the last years in NLP4RE research~\cite{zhao2021natural}.

However, several other RE tasks have profited from ML and even DL techniques, for example:  glossary extraction, with the usage of unsupervised learning~\cite{arora2016automated} and convolutional neural networks~\cite{winkler2016automatic};  requirements classification with the early works from Casamayor \textit{et al.}~\cite{casamayor2010identification} and developments from Kurtanovic and Maalej ~\cite{kurtanovic2017automatically}; requirements tracing~\cite{guo2017semantically,sultanov2013application}, which can be regarded as the field where ML/DL have been more widely experimented for traditional requirements,  especially due to the inherent nature of the problem, which entails finding relevant relationship (i.e., trace links) within a large amount of potential ones. 

With the advent of DL and transfer learning in particular, initial experiments have been carried out in RE with promising results. In particular, DL-based approaches have been proposed to classify software requirements into FR or NFR \cite{hey2020norbert}, to discovery requirements from open source issue reports \cite{li2020deep} and to extract and classify requirements from software project contracts \cite{gonzalez2020comparing}. We predict that research in developing DL-based approaches for RE tasks will grow rapidly in the coming years, overtaking the work on ML-based approaches.

\section{Method}
\label{sec:method}

The main source of the literature used for our data collection is the set of 404 NLP4RE studies identified in our systematic review \cite{zhao2021natural}\footnote{The references of these papers are made available by Zhao \textit{et al.} at: https://github.com/waadalhoshan/NLP4RE.}, which covers the studies up to 2019. We then performed a complementary targeted review to identify recent publications, to find more recent techniques emerging in the last 2 years. This complementary review focused on the major RE and software engineering conferences (i.e., RE, REFSQ and ICSE) and journals (i.e., REJ, JSS, ASE, DKE, IST, and TSE). Based on this updated literature, we extract NLP techniques.

To help us identify and extract NLP techniques from each paper, we followed this definition: "\textit{An NLP technique is a practical method, approach, process, or procedure for performing a particular NLP task, such as POS tagging, parsing or tokenizing} \cite{zhao2021natural}." Our data extraction resulted in a large collection of diverse terms and phrases. To synthesize different terms and phrases into a coherent set of standard terms, we consulted many books written by NLP experts (e.g., \cite{jurafsky2000speech,bird2009natural,sarkar2016text}). This process gave rise to a total of \textbf{57 different NLP techniques}. Table \ref{tab:nlp-tech1} and Table \ref{tab:nlp-tech2} summarize these 57 techniques.

\begin{table*}[ht!]
\centering
\caption{List of NLP Techniques (Part 1)}
\label{tab:nlp-tech1}

\begin{tabular}{p{0.4cm}|p{4cm}|p{11cm}}
\hline

\textbf{ID} &\textbf{Name} & \textbf{Explanation} \\ \hline

1 &
  Part-of-Speech (POS) Tagging &
  POS Tagging (or Tagging) processes a sequence of words, and attaches a POS tag to each word. Parts of speech are also known as word classes or lexical categories. \\ \hline
2 &
  Term Extraction &
  The process of extracting the most relevant words and expressions from text. Related terms: Keyword Extraction, Word Extraction\\ \hline
3 &
  Keyword Searching &
  The technique of finding strings that match a pattern. Related terms: Term Matching, Word Matching \\ \hline
4 &
  Chunking &
  Chunking (or text chunking) is a type of shallow parsing that analyses a sentence by first identifying its constituent parts (nouns, verbs, adjectives, etc.) and then links them to higher order units that have discrete grammatical meanings (noun groups or phrases, verb groups, etc.). Related term: Shallow Parsing. \\ \hline
5 &
  Named Entity Recognition (NER) &
  Subtask of information extraction that is based to find and classify named entities in a certain text into pre-defined categories or class such as the names of persons, organizations, locations, etc. Related terms: Entity Identification, Concept Extraction. \\ \hline
6 &
  Semantic Role Labelling (SRL) &
  The process of detecting the semantic arguments linked with the predicate or verb of a sentence and their classification into their specific roles. Related Term: Semantic parsing, semantic trees, shallow parsing, and shallow semantic analysis. \\ \hline
7 &
  Temporal Tagging &
  The task of finding phrases with temporal meaning within the context of a larger document. \\ \hline
8 &
  Dependency Parsing &
  Dependency parsing is the process of analyzing the grammatical structure of a sentence based on the dependencies between the words in a sentence. Related terms: Syntactic Patterns, Syntactic Structure \\ \hline
9 &
  Constituency Parsing &
  The process of analyzing the sentences by breaking down it into sub-phrases also known as constituents. These sub-phrases belong to a specific category of grammar like NP (noun phrase) and VP(verb phrase). Related terms: Phrase Parsing, Phrase Detection, Phrasal Verb Extraction \\ \hline
10 &
  Link Grammar &
  Builds relations between pairs of words, rather than constructing constituents in a phrase structure hierarchy. \\ \hline
11 &
  Semantic Parsing &
  The task of converting a natural language utterance to a logical form: a machine-understandable representation of its meaning. \\ \hline
12 &
  Sentiment Analysis &
  The process of computationally identifying and categorizing opinions expressed in a piece of text \\ \hline
13 &
  Text Annotation &
  The practice and the result of adding a note or gloss to a text, which may include highlights or underlining, comments, footnotes, tags, and links. \\ \hline
14 &
  Semantic Annotation &
  The process of attaching to a text document or other unstructured content, metadata about concepts (e.g., people, places, organizations, products or topics) relevant to it. \\ \hline
15 &
  Topic Modelling &
  A type of statistical model for discovering the abstract "topics" that occur in a collection of documents \\ \hline
16 &
  Summarization &
  The practice of breaking down long publications into manageable paragraphs or sentences. The procedure extracts important information while also ensuring that the paragraph's sense is preserved. \\ \hline
17 &
  Latent Dirichlet Allocation (LDA) &
  The process of analysing relationships between a set of documents and the terms they contain by producing a set of concepts related to the documents and terms. \\ \hline
18 &
  Latent Semantic Indexing (LSI) &
  A mathematical practice that helps classify and retrieve information on particular key terms and concepts using singular value decomposition (SVD). Related Term: Latent Semantic Analysis (LSA) \\ \hline
19 &
  Semantic Patterns &
  Semantic patterns are generated based on common matching concepts. The top matching concepts of each word are considered. One semantic pattern can relate to several concepts and a single semantic clique can contain several semantic patterns. \\ \hline
20 &
  Case Grammar &
  A system of linguistic analysis, focusing on the link between the valence, or number of subjects, objects, etc., of a verb and the grammatical context it requires. \\ \hline
21 &
  Semantic Frames &
  A coherent structure of concepts that are related such that without knowledge of all of them, one does not have complete knowledge of one of the either. \\ \hline
22 &
  Knowledge Graph &
  A way of storing data that resulted from an information extraction task. \\ \hline
23 &
  Bag-of-Words (BOW) &
  A representation that turns arbitrary text into fixed-length vectors by counting how many times each word appears. This process is often referred to as vectorization. \\ \hline
24 &
  Word Frequency &
  How often a word appears in a document, divided by how many words there are. Related Terms: Term Frequency, Domain Term Frequency \\ \hline
25 &
  Term Frequency-Inverse Document Frequency (TF-IDF) &
  A statistical measure that evaluates how relevant a word is to a document in a collection of documents. \\ \hline
26 &
  Co-location Analysis &
  A Co-location is an expression consisting of two or more words that correspond to some conventional way of saying things. \\ \hline
27 &
  Term-Document Matrix &
  A mathematical matrix that describes the frequency of terms that occur in a collection of documents. \\ \hline
28 &
  Character Counting &
  Counts the number of characters in a line of text, page or group of text. \\ \hline
29 &
  Concordance &
  An alphabetical list of the words (especially the important ones) present in a text, usually with citations of the passages in which they are found. \\ \hline
30 &
  Cosine Similarity &
  A metric used to measure how similar the documents are irrespective of their size. \\ \hline

\end{tabular}
\end{table*}
\begin{table*}[ht!]
\centering
\caption{List of NLP Techniques (Part 2)}
\label{tab:nlp-tech2}

\begin{tabular}{p{0.4cm}|p{4cm}|p{11cm}}
\hline

\textbf{ID} &\textbf{Name} & \textbf{Explanation} \\ \hline

31 &
  Lexical Affinity &
  Assigns to arbitrary words a probabilistic 'affinity' for a particular category. \\ \hline
32 &
  Similarity Distance &
  Determines the minimum number of single character edits required to change one word to another. \\ \hline
33 &
  Document Similarity &
  Computing the similarity between two text documents by transforming the input documents into real-valued vectors. \\ \hline
34 &
  Lexical Similarity &
  Provides a measure of the similarity of two texts based on the intersection of the word sets of same or different languages. \\ \hline
35 &
  Regular Expression &
  A special series of strings for describing a a text pattern for the purpose of searching or replacing the described items. \\ \hline
36 &
  Lexical Patterns &
  Words or chuck of text that occurs in language with high frequency and the meaning of the parts are sometime different than the meaning of the whole. \\ \hline
37 &
  Generation Rules &
  Generation rules to produce meaningful sentences in Natural Language. \\ \hline
38 &
  Stemming &
  A crude heuristic process that chops off the ends of words in the hope of achieving this goal correctly most of the time, and often includes the removal of derivational affixes. \\ \hline
39 &
  Lemmatization &
  Use a vocabulary and morphological analysis of words, normally aiming to remove inflectional endings only and to return the base or dictionary form of a word, which is known as the lemma. \\ \hline
40 &
  Stop-Word Removal &
  Words which are filtered out before or after processing of natural language data (text). \\ \hline
41 &
  Noise Removal &
  Removing characters digits and pieces of text that can interfere with your text analysis. \\ \hline
42 &
  Punctuation Removal &
  Removing puncuatations marks. \\ \hline
43 &
  Lowercasing &
  Converting all your data to lowercase helps in the process of preprocessing and in later stages in the NLP application, when you are doing parsing. \\ \hline
44 &
  Camel Case Splitting &
  Split CamelCase string to individual strings. \\ \hline
45 &
  Tokenization &
  The process of breaking a stream of text into words, phrases, symbols, or other meaningful tokens. Related terms: Word Segmentation \\ \hline
46 &
  Sentence Segmentation &
  Split a document into sentences, each containing a list of tokens. Related terms: Sentence Splitting \\ \hline
47 &
  n-gram &
  A representation of a text using a sequence of N words or N characters (character n-gram), where N can be any number. Thus we can have 1-gram (unigram), 2-gram (bigram), 3-gram (trigram), etc. \\ \hline
48 &
  Word Embedding &
  One of the most popular technique to learn word embeddings using shallow neural network. Word embeddings are vector representations of a particular word. Related terms: Word2Vec \\ \hline
49 &
  Contextualized word embedding &
  A neural model that learns a generic embedding function for variable length contexts of target words. Related terms: Context2Vec \\ \hline
50 &
  Sentence and document Embedding &
  A generalized word2vec method, for representing documents as a vector. Related term: Doc2Vec \\ \hline
51 &
  GloVe &
  An alternative to word2vec for the representation of the distributed words. \\ \hline
52 &
  FastText &
  An alternative to word2vec, FastText represents each word as a bag of character n-gram. \\ \hline
53 &
  Textual Entailment Recognition &
  Deciding, given two text fragments, whether the meaning of one text is entailed (can be inferred) from another text. \\ \hline
54 &
  Homonym Detection &
  Detecting the words that are pronounced the same as each other (e.g., "maid" and "made") or have the same spelling (e.g., "lead weight" and "to lead"). \\ \hline
55 &
  Synonym Detection &
  Finding a a word or phrase that means exactly or nearly the same as another word or phrase in a text. \\ \hline
56 &
  Coreference Resolution &
  Finding all expressions that refer to the same entity in a discourse. \\ \hline
57 &
  Anaphora Resolution &
  Resolving what a pronoun, or a noun phrase refers to in a discourse. \\ \hline

\end{tabular}
\end{table*}

\section{Classifying NLP Techniques by Tasks}
\label{sec:NLPTasks}

\begin{figure}
  \centering
    \includegraphics[width=8cm]{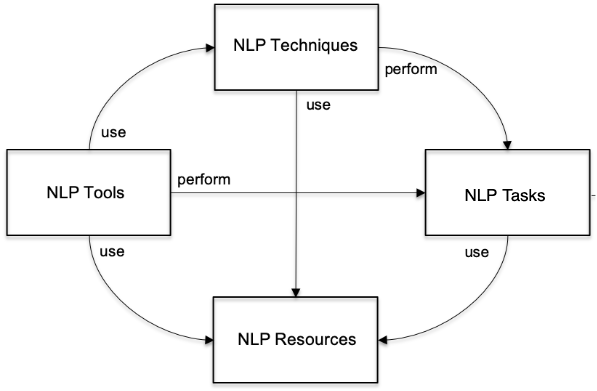}
    \caption{Relationship between NLP techniques and NLP Tasks.}
    \label{fig:NLPTasks}
\end{figure}

We first classify the NLP techniques based on their text processing tasks. Figure \ref{fig:NLPTasks} depicts the relationship between NLP techniques, NLP tasks, NLP resources, and tools. We define \textit{a NLP task as a piece of text processing work that can be done by means of one or more NLP techniques, supported by some NLP tools and resources}. A list of frequently performed NLP tasks in RE are desribed below:

\begin{itemize}
\item \textbf{Part-of-Speech (POS) Tagging:} To associate words with part-of-speech (POS) tags to distinguish between nouns, verbs, adjectives, adverbs, etc. The input unit is a sentence, as context words (i.e., neighbouring ones) are normally used to infer the POS of a word. 

\item \textbf{Semantic Tagging:} To extract useful bits of information (words, terms, relations, etc.) from the text.

\item \textbf{Syntactic Analysis:} To analyze the syntactic structure of a sentence to represent the relationship between its components. Different representation structures can be used, such as the parse tree, or the dependency parsing graph. 

\item \textbf{Semantic Analysis:} To identify and label semantically relevant components and relations in the text. These entails identifying the meaning of a certain word or phrase in a context and the relationship between words or terms.

\item \textbf{Frequency Analysis:} To analyze the frequencies of words or terms in a certain context and to produce probabilistic data.

\item \textbf{Similarity Analysis:} To calculate the numerical estimates of similarity between text elements, for example to identify semantic relatedness, synonyms, or to support topic modelling. 

\item \textbf{Rule-Based Analysis:} To use grammar rules, semantic rules or patterns to analyse the syntax of a text.

\item \textbf{Text Normalization:} To convert the words into their original form and remove
unnecessary words or characters from the text. 

\item \textbf{Text Segmentation:} To break down a text into a sequence of individual sentences or words. 

\item \textbf{Text Normalization:} To reduce the words to a standardised format, with the removal of stop words, and reduction of typographical forms (e.g., upper case, camel case) to a unique form.

\end{itemize}

\begin{table*}[ht!]
\centering
\caption{Classifying NLP Techniques by Tasks.}
\label{tab:nlp-tasks}

\begin{tabular}{p{3cm}|p{5.5cm}|p{7.5cm}}
\hline

\textbf{NLP Tasks} &\textbf{Explanation} & \textbf{NLP Techniques} \\ \hline
  
\rule{0pt}{3ex}\textbf{Part-of-Speech Tagging} & Associate words with part-of-speech (POS) tags to distinguish between nouns, verbs, adjectives, adverbs, etc. &
 Part-of-Speech (POS) Tagging \\ \hline
  
\rule{0pt}{3ex}\textbf{Semantic Tagging} &
  Extract useful bits of information (words, terms, relations, etc.) from the text. &
  Term Extraction, Term Matching, Chunking, Concept Extraction, Named Entity Recognition (NER), Semantic Role Labelling (SRL), Temporal Tagging \\ \hline
  
\rule{0pt}{3ex}\textbf{Syntactic Analysis} &
Construct a syntactic structure representing the relationship between the logical components in a stream of text, such as the parse tree, or the dependency parsing graph. &
 Dependency Parsing, Constituency Parsing, Link Grammar \\ \hline
  
\rule{0pt}{3ex}\textbf{Semantic Analysis} &
 Identify and label semantically relevant components and relations in the text. &
 Semantic Parsing,  Sentiment Analysis, Text Annotation, Semantic Annotation, Topic Modelling, Summarization, Latent Dirichlet Allocation (LDA), Latent Semantic Indexing (LSI),  Semantic Patterns, Case Grammar, Semantic Frames, Knowledge Graph, Textual Entailment Recognition (TER), Homonym Detection, Synonym Detection, Coreference Resolution, Anaphora Resolution\\ \hline
  
\rule{0pt}{3ex}\textbf{Frequency Analysis} &
  Analyse the frequency of occurrence of lexical elements (e.g., words and characters) and groups of elements (e.g., phrases  and multiwords) in a given text. &
  Bag-of-Words (BOW), Word Frequency, Term Frequency (TF), Term Frequency \& Inverse Document Frequency (TF-IDF), Co-location Analysis, Term-Document Matrix, Character Counting, Concordance \\ \hline
  
\rule{0pt}{3ex}\textbf{Similarity Analysis} &
Calculate numerical values of the similarity between text elements, such as to identify semantic relatedness. &
Cosine Similarity,  Lexical Affinity, Similarity Distance,  Document Similarity, Lexical Similarity\\ \hline
  
  \rule{0pt}{3ex}\textbf{Rule-Based Analysis} &
 Use rules or patterns to analyse the syntax or semantics of a text or transform the text. &
 Regular Expression, Lexical Patterns, Generation Rules \\ \hline
  
\rule{0pt}{3ex}\textbf{Text Normalization} &
  Convert the words into their original form and remove unnecessary  words or characters from the text. &
  Stemming, Lemmatization, Stop-Word Removal, Noise Removal, Punctuation Removal,  Lowercasing, Camel Case Splitting \\ \hline
  
\rule{0pt}{3ex}\textbf{Text Segmentation} &
 Break down a text into a sequence of individual tokens (i.e., words or sentences). &
 Tokenization,  Sentence Segmentation \\ \hline
  
\rule{0pt}{3ex}\textbf{Text Representation} &
  Represent words, sentences or documents using vectors of real numbers. &
  N-gram, Word2Vec, Context2Vec, Doc2Vec, GloVe, FastText\\ \hline
  
\end{tabular}

\end{table*}

Table \ref{tab:nlp-tasks} presents the classification results of the NLP techniques for RE based on these tasks.

\section{Classifying NLP Techniques by Linguistic Analysis Levels}
\label{sec:NLPLevels}

\begin{table*}[ht!]

\caption{Classifying NLP Techniques by Levels of Analysis.}
\label{tab:nlp-levels}

\begin{tabular}{p{2cm}|p{7.5cm}|p{7cm}}
\hline

\textbf{Analysis Level} &\textbf{Explanation} & \textbf{NLP Techniques} \\ \hline

\rule{0pt}{3ex}\textbf{Morphology} & This is the lowest level of text analysis, dealing with the smallest parts of words that carry meaning. All the techniques used for text normalization belong to this category. &
Stemming, Lemmatization, Stop-Word Removal, Noise Removal, Punctuation Removal, Lowercasing, Camel Case Splitting \\ \hline

\rule{0pt}{3ex}\textbf{Lexical} & This is the word-level of text analysis, interpreting the meaning of individual words to gain word-level understanding. All the techniques used for frequency analysis belong to this category. In addition, Tokenization and n-gram should also be in this category. &
BOW, TF, TF-IDF, Co-location Analysis, Term-Document Matrix, Character Counting, Concordance, n-gram \\ \hline

\rule{0pt}{3ex}\textbf{Syntactic} & This level focuses on analyzing
the words in a sentence through the grammatical structure of the sentence. All the techniques used for syntactic analysis belong to this category. In addition, the techniques used for text segmentation and Regular Expression for Rule-Based Analysis should also belong to this category. &
POS Tagging, Dependency Parsing, Constituency Parsing,  Link Grammar, Regular Expression, Tokenization, Sentence Segmentation \\ \hline

\rule{0pt}{3ex}\begin{flushleft}\textbf{Semantic (Word-Level)}\end{flushleft}& We split the semantic level into word-level semantic and sentence-level semantic. The word-level semantic focuses on the meanings of individual words (e.g., dictionary definitions of words and word-sense disambiguation). Most techniques used for semantic tagging and similarly analysis belong to this level. In addition, apart from n-gram, the techniques used for text representation belong to this category. &
Term Extraction, Keyword Searching, Chunking,  NER, Temporal Tagging, Lexical Patterns, Cosine Similarity, Lexical Affinity, Similarity Distance,  Document Similarity, Lexical Similarity, Word2Vec, Context2Vec, Doc2Vec, GloVe, FastText\\ \hline
  
\rule{0pt}{3ex}\begin{flushleft}\textbf{Semantic (Sentence-Level)}\end{flushleft} & This level deals with the compositional semantics, which looks at the interactions among word-level meanings in sentences (e.g., semantic role labeling). Most techniques used for semantic analysis belong to this category. In addition, SRL and most techniques for disambiguation should also belong to this category.&
Semantic Parsing, Sentiment Analysis, Text Annotation, Semantic Annotation, Topic Modelling, SRL, Summarization, LDA, LSI, Semantic Patterns, Case Grammar, Semantic Frames, Knowledge Graph, TER, Homonym Detection, Synonymy Detection \\ \hline
 
 \rule{0pt}{3ex}\textbf{Discourse} & This level focuses on the properties of the text as a whole that convey meaning by making connections between component sentences. Only three techniques belong to this category. &
Coreference Resolution, Anaphora Resolution, Generation Rules \\ \hline

\end{tabular}

\end{table*}

Here, we classify the NLP techniques by levels of linguistic analysis. According to Liddy \cite{liddy2001natural}, linguistic analysis can be performed at the following seven levels:

\textbf{Phonology}. This level deals with the interpretation of speech sounds within and across words.

\textbf{Morphology.} This is the lowest level of text analysis. At this level, a NLP technique analyzes the smallest parts of words that carry meaning, which are composed of morphemes, including prefixes, roots and suffixes of words.

\textbf{Lexical.} A NLP technique at this level can interpret the meaning of individual words to gain \textit{word-level understanding} \cite{liddy2001natural}. 
Lexical analysis may require a lexicon or dictionary, which may be quite simple, with only the words and their POS tags, or may be increasingly complex and contain information on the semantic class of the word, its arguments etc. \cite{liddy2001natural}.  

\textbf{Syntactic.} A NLP technique at this level focuses on analyzing the words in a sentence through the grammatical structure of the sentence. This requires both a grammar and a parser \cite{liddy2001natural}. There are two general types of parser: dependency and constituency \cite{jurafsky2000speech}. The dependency parser produces a syntactic representation of a sentence based on the dependencies between the words in the sentence, whereas the constituency parser represents a sentence as a parse tree of related constituents (i.e., sub-phrases). These representations (i.e., syntax) carry meaning in most languages, because the the arrangement of words or sub-phrases in a sentence contributes to meaning \cite{liddy2001natural}. 

\textbf{Semantic.} A NLP technique at this level may focus on the meanings of individual words (e.g., dictionary definitions of words and word-sense disambiguation), or compositional semantics, which looks at the interactions among word-level meanings in sentences (e.g., semantic role labeling). Semantic analysis thus can be divided into \textit{word-level semantic} and \textit{sentence-level semantic} (groups of words or sentence-level). Semantic role labelling  \cite{gildea2002automatic} and Case Grammar \cite{fillmore1976frame,fillmore2003framenet} are among the examples of semantic analysis techniques.

\textbf{Discourse.} A NLP technique at this level focuses on the properties of the text as a whole that convey meaning by making connections between component sentences. Several types of discourse processing can occur at this level, two of the most common being \textit{anaphora resolution} and \textit{coreference resolution} \cite{liddy2001natural}. 

\textbf{Pragmatic.} This is the highest level of NLP. To reach this level, NLP techniques need to be able to achieve human-like language understanding, the ultimate goal of natural language understanding (NLU). This entails inferring extra meaning from texts that is not actually encoded in them \cite{liddy2001natural} and understanding narratives according to different contexts and with respect to different actors and their intentions \cite{cambria2014jumping}. This requires NLP tools to have world knowledge and human intelligence, and the ability to project semantics and sentics dynamically \cite{cambria2014jumping}. Pragmatic analysis appears to be the most challenging NLP curve to jump \cite{cambria2014jumping}.

It is assumed that humans normally produce or comprehend language by utilizing all of these levels \cite{chowdhury2003natural}. These levels thus represent the competence of a NLP tool: The more levels of analysis the tool supports, the stronger or more capable the tool; the more higher-levels of analysis the tool supports, the more advance the tool. 

Table \ref{tab:nlp-levels} classifies the NLP techniques based on these levels. As the table shows, we have not found any techniques for the phonetic level analysis, as NLP techniques have been largely use to deal with texts (including requirements documents \cite{ferrari2017pure}, app reviews, tweets, social media posts and usage data~\cite{ferrari2017natural,dalpiaz2018natural,ferrarinlp,maalej2015toward,maalej2019data}, legal documents~\cite{sleimi2018automated}, and privacy policies~\cite{bhatia2016mining}). In addition, we have not found any techniques for pragmatic analysis either. This is because NLP4RE research has so far focused on text processing of documents and has not reached the level of natural language understanding (NLU).

\section{Conclusion}
\label{sec:conclusion}

This paper presents 57 commonly used NLP techniques in RE and organizes them in two different ways: by their NLP tasks and by their analysis levels. The organization provides a knowledge base for sharing these techniques. A user of this knowledge base can query each NLP technique progressively: Through Table \ref{tab:nlp-tech1} and Table \ref{tab:nlp-tech2}, the user can ask: What technique is it? Does it work at the word-level or sentence-level of text processing? Based on Table \ref{tab:nlp-tasks}, the user can ask: Which text processing task does this technique support? What are the alternative techniques for the same task? From Table \ref{tab:nlp-levels}, the user can ask: What level of language analysis does this technique provide? What are the techniques for performing other levels of analysis? The answers to these questions can help the user to decide if a specific technique is relevant to the task at hand. 

Our future work will improve this knowledge base as follows:

\begin{itemize}
\item To show the relationship between a given technique and other techniques. For example, for text normalization, what techniques can I use together? in what order? For text representation, which technique is better for my case?
\item To provide information on the available NLP tools that support each technique.
\end{itemize}

\nocite{*} 
\bibliographystyle{IEEEtran}
\bibliography{Bibliography}

\end{document}